\documentclass{article}

\usepackage{arxiv}

\usepackage[utf8]{inputenc} 
\usepackage[T1]{fontenc}
\usepackage{hyperref} 
\usepackage{url}   
\usepackage{booktabs}      
\usepackage{amsfonts}      
\usepackage{nicefrac}     
\usepackage{microtype}    
\usepackage{lipsum}		
\usepackage{graphicx}
\usepackage{natbib}
\usepackage{doi}

\usepackage{xcolor}      
\usepackage{bm}
\usepackage{enumerate}
\usepackage{enumitem}

\usepackage{amsmath}
\usepackage{amssymb}
\usepackage{mathtools}
\usepackage{amsthm}

\usepackage{bbding}
\usepackage{pifont}
\usepackage{color,xcolor}
\usepackage{wasysym}

\usepackage{color}
\usepackage{subfigure}
\usepackage{booktabs}

\usepackage{multirow}
\usepackage{multicol}
\usepackage{hyphenat}

\usepackage{amsmath}
\usepackage{float}
\usepackage{subfigure}
\usepackage{makecell}
\usepackage{enumitem}
\usepackage[ruled,boxed,linesnumbered]{algorithm2e}

\newtheorem{theorem}{Theorem}

\usepackage[capitalize,noabbrev]{cleveref}
\usepackage{wrapfig}

\def\mb{\mathbf}
\def\mc{\mathcal}

\graphicspath{{figs/}}

\title{Towards Stable Training of Parallel Continual Learning}

\author{
        Yuepan Li$^1$, Fan Lyu$^2$, Yuyang Li$^3$, Wei Feng$^1$, Guangcan Liu$^3$, Fanhua Shang$^{2}$\thanks{Corresponding author.}\\\\
        $^1$ College of Intelligence and Computing, Tianjin University\\
        $^2$ New Laboratory of Pattern Recognition, Institute of Automation, Chinese Academy of Sciences\\
        $^3$ School of Automation, Southeast University\\
        \texttt{\{yuepanli, fhshang\}@tju.edu.cn},\quad\texttt{fan.lyu@cripac.ia.ac.cn},\quad\texttt{yuyangli@seu.edu.cn}\\
        \texttt{wfeng@ieee.org},\quad\texttt{gcliu1982@gmail.com}
}

\begin{document}

\maketitle

\begin{abstract}
  Parallel Continual Learning (PCL) tasks investigate the training methods for continual learning with multi-source input, where data from different tasks are learned as they arrive.
  PCL offers high training efficiency and is well-suited for complex multi-source data systems, such as autonomous vehicles equipped with multiple sensors.
  However, at any time, multiple tasks need to be trained simultaneously, leading to severe training instability in PCL. 
  This instability manifests during both forward and backward propagation, where features are entangled and gradients are conflict.
  This paper introduces Stable Parallel Continual Learning (SPCL), a novel approach that enhances the training stability of PCL for both forward and backward propagation. 
    For the forward propagation, we apply Doubly-block Toeplit (DBT) Matrix based orthogonality constraints to network parameters to ensure stable and consistent propagation. 
    For the backward propagation, we employ orthogonal decomposition for gradient management stabilizes backpropagation and mitigates gradient conflicts across tasks. 
  By optimizing gradients by ensuring orthogonality and minimizing the condition number, SPCL effectively stabilizing the gradient descent in complex optimization tasks. 
  Experimental results demonstrate that SPCL outperforms state-of-the-art methjods and achieve better training stability. The source code is publicly available at \url{https://github.com/jasmine-0/SPCL}.
\end{abstract}

\section{Introduction}

In the era of Internet of Things (IoT) and embodied artificial intelligence, effectively utilizing multi-source data to train multi-task models becomes important.
For example in applications including autonomous driving~\cite{shaheen2022continual} and monitoring systems~\cite{doshi2020continual}, they rely on the perception from multiple sensors to improve the decision.
When multi-source data become dynamic over long periods, experiencing increments due to external factors, model training becomes dynamic as well.
This learning task is referred to as Parallel Continual Learning (PCL)~\cite{lyu2023measuring}.
In contrast to traditional Continual Learning (CL)~\cite{parisi2019continual, de2021continual}, which requires models to learn new tasks one by one, PCL allows multi-task training at any time.
PCL can continuously adapt to new environmental variables or changing conditions without requiring system downtime or pending.

In traditional CL, catastrophic forgetting~\cite{kirkpatrick2017overcoming} happens when a model learns new information and unintentionally erases previous knowledge.
More than catastrophic forgetting, PCL faces significant training conflict challenge~\cite{lyu2023measuring}.
Training conflicts arise because different tasks inherently exhibit significant differences and varied training progress. This discrepancy often leads to conflicting multi-task gradients, resulting in unstable model updates.
To solve the challenges, Lyu \textit{et al.}~\cite{lyu2023measuring} propose MaxDO, which uses asymmetric gradient distance metric and minimizes gradient discrepancies among tasks, addressing instability and interference issues in feature extractors.
However, while MaxDO reduces training conflicts and catastrophic forgetting, its performance can still be susceptible to the variations in task relevance and the scale of gradient changes across different tasks, which may destabilize the training process.

In this paper, we delve deeper into the causes of training instability in PCL. 
{Inspired by the stable training of CNNs}~\cite{eilertsen2019single,huang2018orthogonal,lezcano2019cheap,wang2020orthogonal}, we present that the instability in PCL training arises because both forward and backward propagation introduce unstable factors.
For the forward propagation, learning multiple tasks simultaneously with shared network layers leads to interference among feature extractors, further compromising task-specific knowledge acquisition, and destabilizing the training process.
For the backward propagation, PCL models exhibit instability due to conflicting learning objectives that can disrupt optimization and destabilize weight updates during gradient descent.

Motivated by this, we propose to improve the training stablility of PCL, and present a Stable Parallel Continual Learning (SPCL) method based on orthogonalization strategy. 
We conduct a theoretical analysis on the causes of training instability in PCL models, identifying the \textit{condition number} as a key metric for assessing gradient system stability.
First, we integrate Doubly-block Toeplitz matrix based orthogonality into the PCL framework, applying it to convolutional layers for stable forward propagation.
Then, we employ orthogonal decomposition for improved backpropagation stability. This method effectively addresses gradient conflicts, enhancing accuracy, convergence, and retention in dynamic learning environments. 
In our experiments, our methods have yielded impressive results on the EMNIST, CIFAR-100, and ImageNet datasets, significantly advancing PCL models and providing a robust solution to training instability.
Our contributions are three-fold:

\begin{enumerate}[label=(\arabic*),left=0pt,itemsep=0pt]
    \vspace{-5px}
    \item To the best of our knowledge, we are the first to analyze the cause of training instability in PCL, which lies in the introduction of unstable factors during both forward and backward propagation. Based on this finding, we propose the SPCL framework.
    \item We apply orthogonality constraints to CNN filters, which decreases feature redundancy and prevents gradient explosion and vanishing, enhancing the model's reliability.
    \item We adjust the update by orthogonally aligning the principal components of the gradient matrix and introducing a relaxation term. This combined approach reduces conflicts among gradients, maintains scale, and lowers the condition number, enhancing the stability of the training process.
\end{enumerate}

\section{Related Work}

\textbf{CL and PCL.}
CL enables models to continuously learn new information while retaining old knowledge, crucial for intelligent systems in dynamic environments~\cite{parisi2019continual}. A primary challenge of CL is catastrophic forgetting~\cite{kirkpatrick2017overcoming}, where CL models lose previous knowledge when learning new tasks, highlighting traditional machine learning's limitations in multitasking and adaptive scenarios.
To address catastrophic forgetting, key strategies in existing research include regularization-based methods, rehearsal-based methods, and architecture-based methods~\cite{de2021continual}. Regularization methods~\cite{kirkpatrick2017overcoming,aljundi2018memory,chaudhry2018riemannian,du2022agcn}, like Elastic Weight Consolidation (EWC)~\cite{kirkpatrick2017overcoming}, incorporate constraints within the loss function to limit forgetting. Replay methods~\cite{lopez2017gradient,shah2018distillation,sun2022exploring,liu2023centroid}, such as Gradient Episodic Memory (GEM)~\cite{lopez2017gradient}, retain old memories by preserving old data or creating surrogate data. Architecture-based methods~\cite{rusu2016progressive,yoon2017lifelong} allocate distinct parameters to each task to prevent interference. 
Traditional CL typically focuses on sequentially learning tasks, which may lead to significant delays in integrating new information for multi-source tasks.
The PCL task addresses this issue. Lyu et al.~\cite{lyu2023measuring} proposed measuring asymmetric distances between gradients to obtain optimal update. However, the gradients obtained by this method neglect training stability during both forward and backward propagation.

\textbf{Weight Orthogonal Regularization.}
Incorporating orthogonality constraints in CNNs has proven to enhance their performance. 
In practice, orthogonal regularization improves CNN performance in tasks like image recognition, object detection, and semantic segmentation~\cite{vu2019advent}. Early studies indicates~\cite{saxe2013exact} that applying orthogonal initialization at the start of training significantly boosts efficiency and overall network performance.
Existing research shows that applying orthogonality constraints during neural network training can significantly enhance model performance and stability~\cite{xie2017all,huang2018orthogonal,bansal2018can,lezcano2019cheap,wang2020orthogonal}. 
Two kinds of orthogonality constraints~\cite{achour2022existence} are well-studied.
\textit{Hard orthogonality}~\cite{jia2017improving,huang2018orthogonal}, which keeps weight matrices strictly orthogonal throughout training using techniques like Stiefel manifold representation. 
\textit{Soft orthogonality}~\cite{balestriero2018spline,bansal2018can,balestriero2020mad}, which adds a regularization term $ \| \mb{W}^\top \mb{W} - \mb{I} \|_2 $ to the loss function.
Kernel orthogonality is also important in CNNs~\cite{brock2018large}, which ensures that each convolutional kernel, treated as a matrix, has linearly independent rows or columns, while convolutional layer orthogonality views the entire layer as a composite operation that maps input to output feature space, maintaining orthogonality between the two.

\textbf{Gradient Manipulation.}
In PCL, managing multiple gradients is critical. 
Due to the significant differences between new and old task gradients, conflicts often arise, necessitating advanced management methods. 
Gradient reweighting techniques modify the contribution of each task to the overall gradient by assigning varying weights to their loss functions, dynamically based on task uncertainty, helping balance gradients and reduce conflicts~\cite{sener2018multi,fernando2021dynamically,groenendijk2021multi,liu2021conflict}. 
Methods like PCGrad~\cite{yu2020gradient} and CAGrad~\cite{liu2021conflict} focus on minimizing gradient conflicts through orthogonal projections and optimizing update vectors for equitable learning across tasks, respectively.
Additionally, Aligned-MTL~\cite{senushkin2023independent} uses SVD to align task gradients, reducing directional discrepancies.
The OGD~\cite{farajtabar2020orthogonal} method addresses catastrophic forgetting by orthogonally projecting new task gradients to previous tasks' gradients, minimizing disruption to learned knowledge while integrating new information.
Layerwise Gradient Decompsition~\cite{tang2021layerwise} employs gradient decomposition to separate gradients into shared and task-specific components, promoting updates that align with shared gradients and are orthogonal to task-specific ones, effectively reducing task interference and preserving unique task knowledge.

\section{Stable Training in PCL}

\subsection{Preliminary: Parallel Continual Learning}

In PCL, $T$ tasks are represented by parallel data streams ${\mathcal{D}_1, \cdots, \mathcal{D}_T}$, each consisting of data tuples $(\mb{x}, \mb{y})$ with $\mb{x}$ as the input and $\mb{y}$ as a one-hot encoded label. 
Unlike traditional CL, which processes tasks sequentially, PCL allows for simultaneous access to all tasks with no fixed training endpoint.
This requires rapid model adaptation while preserving knowledge in diverse tasks. The PCL framework employs a task-neutral backbone with parameters $\boldsymbol{\theta}$, along with a growing set of task-specific classifiers $\boldsymbol{\theta}_t$ for each task. Training in PCL is a dynamic Multi-Objective Optimization (MOO) challenge.
At any time, the number of objectives depends on the number of intraining tasks, and PCL aims at concurrently minimizing empirical risks across tasks by optimizing both shared $\boldsymbol{\theta}$ and task-specific $\{\boldsymbol{\theta}_i | i \in \mc{T}_t\}$ parameters through task-specific loss functions $\ell_i$,
\begin{equation}
    \min \limits_{\boldsymbol{\theta}, \{ \boldsymbol{\theta}_{t} | i \in T_{i}\}} \{ \ell_{i}(\mathcal{D}_{i}) | i \in \mc{T}_t  \},
\end{equation}
where $\mc{T}_t$ is the set of tasks that are active at a given time $t$. The update for the shared parameters $\bm{\theta}$ is influenced by gradients from all currently active tasks. 
Thus, at any time of PCL training, we have multiple gradients $\{\mb{g}_{1}, \cdots, \mb{g}_{|\mc{T}_t|}\}$ for multiple tasks.
PCL navigates a complex optimization landscape filled with local minima and saddle points, which can destabilize updates to $\bm{\theta}$. Additionally, managing conflicts and dominance among multiple task gradients further complicates gradient stability across tasks~\cite{yu2020gradient}. This complexity, combined with the continuous task shifts, poses significant challenges in balancing multiple learning objectives. Following \cite{lyu2023measuring}, our method employs replay based techniques in continual learning, periodically reintroducing old data to prevent catastrophic forgetting while accommodating new information.

\subsection{Stable Training for PCL in perspective of orthogonality}
\label{sec:3_2}

\begin{figure}[t]
\centering
\subfigure[Condition Number]{\centering
  \includegraphics[width=0.245\textwidth]{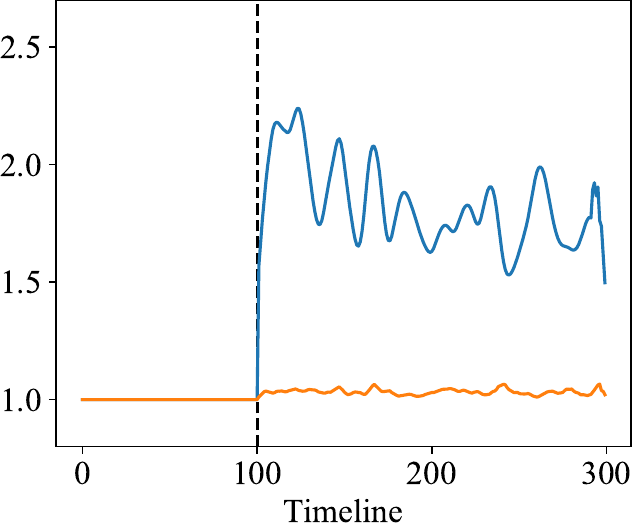}}
\subfigure[Gradient Conflicts]{\centering
  \includegraphics[width=0.245\textwidth]{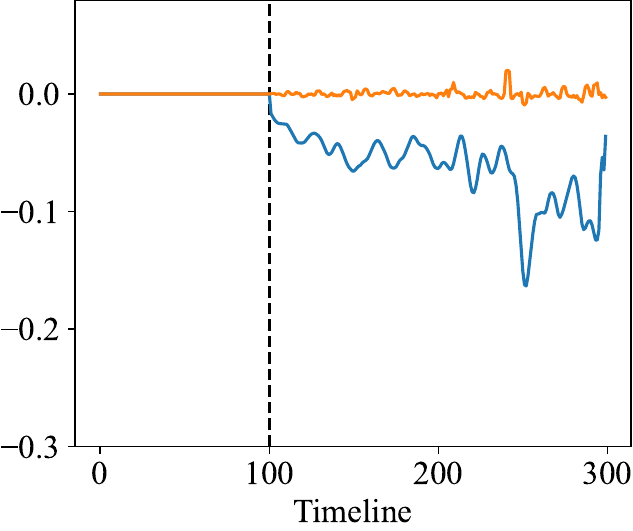}}
\subfigure[Gradient Divergence]{\centering
  \includegraphics[width=0.245\textwidth]{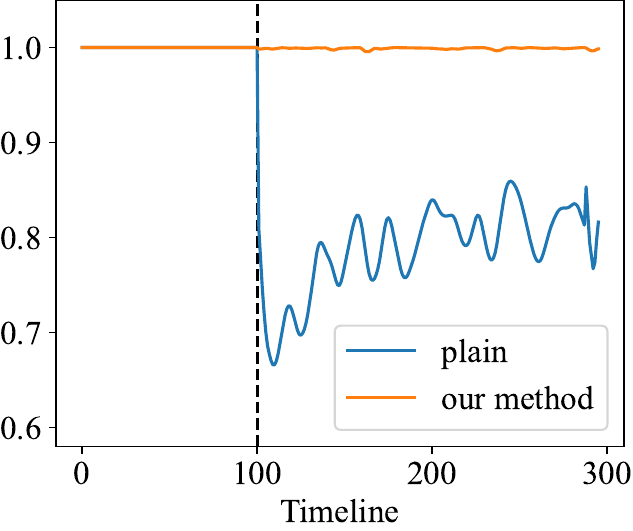}}
\subfigure[Accuracy]{\centering
  \includegraphics[width=0.245\textwidth]{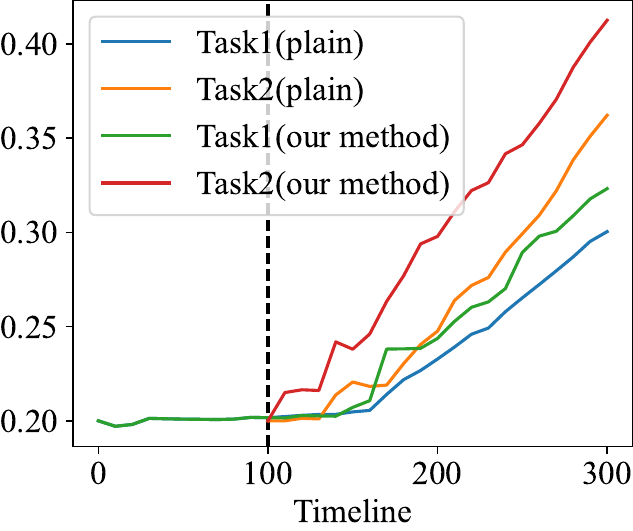}}
\caption{A toy experiment on CIFAR-10, employing a two-task PCL task. During the training process, one task is trained throughout, and in the line graph, a dashed line indicates the point at which the second task joins the training. 
}
\label{fig:stable}
\end{figure}

In PCL, tasks have unique data distributions, causing training instabilities from gradient interference and conflicting parameter updates. This may worsen gradient vanishing or exploding if layer outputs are too small or large, disrupting training and hindering robust feature learning.
To address this challenges, it is crucial to maintain consistent input-output relationships, ensuring smooth transmission of feature signals and preserving the key representations. Moreover, addressing gradient conflicts is essential to harmonize the learning updates across different tasks in PCL, ensuring that the learning from one task does not detrimentally affect the learning of another. 

\textbf{Orthogonalization} constraints within neural networks have been proven to significantly enhance training stability by enforcing orthogonality on linear transformations in hidden layers~\cite{bansal2018can}. 
The preservation of orthogonality in linear transformations can be formalized as follows~\cite{huang2020controllable}.
\begin{theorem} \text{(Preservation of Orthogonality in Linear Transformations.)}
For a linear transformation $\mb{Y} = \mb{K} \cdot \mb{X}$ in CNNs, where $\mb{K}$ is orthogonal ($\mb{K}^\top \mb{K} = I$), the following properties hold:\\
(1) The norm of the output vector $\mb{Y}$ is equal to the norm of the input vector $\mb{X}$, $\| \mb{Y} \| = \| \mb{X} \|$.
(2) The output vector $\mb{Y}$ and its gradient $\frac{\partial{\mathcal{L}}}{\partial{\mb{Y}}}$ preserve the same covariance properties as the input vector $\mb{X}$ and its gradient $\frac{\partial{\mathcal{L}}}{\partial{\mb{X}}}$, respectively.
\label{thm:orth}
\end{theorem}
Theorem~\ref{thm:orth} shows that orthogonalization preserves norms of activations and gradients across layers in PCL, and ensures smoother convergence, thus preventing exploding or vanishing gradients.

To show the importance of orthogonality in PCL, we conduct toy experiments with a two-task PCL model. 
In the toy experiment, the first task is trained first, and the second task is introduced at a specific point. 
During PCL model training, gradient matrices from new tasks $\mb{g}_{2}$ and old tasks $\mb{g}_{1}$ form a combined gradient matrix $\mb{G} = [\mb{g}_{1}, \mb{g}_{2}]$. 
We use three metrics to evaluate the training stability in PCL including the condition number $\kappa$ of $\mb{G}$, cosine distance between the two tasks, and gradient magnitude similarity~\cite{yu2020gradient}, they can be computed as follows. 
\begin{equation}
\kappa(\mb{G}) = \frac{\sigma_{\text{max}}} {\sigma_{\text{min}}},\quad \cos(\mb{g}_1, \mb{g}_2) = \frac{\mb{g}_1 \cdot \mb{g}_2}{\|\mb{g}_1\| \|\mb{g}_2\|},\quad \text{mag}(\mb{g}_1, \mb{g}_2) = \frac{2\|\mb{g}_1\|_2 \|\mb{g}_2\|_2}{\|\mb{g}_1\|_2^2+ \|\mb{g}_2\|_2^2}.
\label{eq:condition_number}
\end{equation}
The results are shown in Figs.~\ref{fig:stable}(a), (b), and (c).
When two tasks are trained simultaneously using the plain approach, values often deviate from optimal levels and display severe fluctuations as training progresses. This significant numerical instability can hinder learning across tasks, as the approach struggles to maintain stability in performance metrics.
The toy experiment shows that instability in PCL arises from challenges in both forward and backward propagation. Issues in signal consistency during forward propagation may cause conflicting activations, while interference among gradients during backward propagation disrupts learning process. These interconnected challenges significantly impact the stability of training in PCL, highlighting the importance of developing targeted strategies to effectively manage these areas.

\section{Method}

\subsection{Overview}

In this paper, we explore the orthogonalization in PCL forward and backward propagation.
During \textit{forward propagation}, overlapping or entangled features may cause conflicting activations. 
In the context of \textit{backward propagation}, the gradient updates from different tasks may conflict. 
On top of this, we present a dual-strategy approach termed SPCL for PCL. 
First, we apply Doubly-block Toeplit (DBT) matrix based orthogonality constraints to network parameters to ensure stable and consistent forward propagation. 
Second, we employ orthogonal decomposition for gradient management stabilizes backpropagation and mitigates gradient conflicts across tasks. 
Both strategies use soft orthogonal regularization under the Frobenius norm, adding minimal computational overhead while enhancing training stability in PCL environments.
These methods collectively enhance PCL model performance by maintaining effective learning across tasks, leading to improved accuracy trends for both tasks, as demonstrated in Fig.~\ref{fig:stable}(d).

\subsection{Doubly-block Toeplit Matrix based Orthogonal Regularization Method}

First, we propose to improve the training stability in the forward propagation.
As mentioned in Sec.~\ref{sec:3_2}, the linear transformation between layers in CNN is computed as $\mb{Y}=\mb{K} \ast \mb{X}$, where $\mb{X}$ and $\mb{Y}$ represent the input and output of each layer, respectively, and $\mb{K}$ denotes the kernel. 
To enhance computational efficiency, we utilize a DBT matrix representation of the kernel (see more details in Appendix~\ref{sec:DBTM}).
From the insights provided in \cite{wang2020orthogonal}, we define the conditions for orthogonality, which is crucial for the orthogonal properties of convolution layers. These conditions are derived from the results of self-convolution computations:
\begin{equation}
\text{Conv}(\mb{K}, \mb{K}, \text{padding} = P, \text{stride} = S) = \mb{I}_{0},
\label{eq:orth_loss_row}
\end{equation}
where $\mb{I}_{0} \in \mathbb{R}^{N \times N \times (2P/S + 1)\times (2P/S + 1)}$ 
is the identity matrix at the centre and zeros entries elsewhere and $k$ is the kernel size. These conditions ensure that the filters within a layer and across layers remain orthogonal, reducing redundancy and improving feature diversity.

To effectively combine the task objectives with the orthogonality requirements, we incorporate a soft orthogonal convolution regularization loss into the overall loss function of backbone network. We assign a positive weight, denoted by $\lambda > 0$, to this orthogonal regularization loss as follows.
\begin{equation}
L = L_{\text{task}} + \alpha \sum\nolimits^{D}_{i=1}  L_{\text{orth}} = L_{\text{task}} + \alpha \sum\nolimits^{D}_{i=1} \|\mb{K}^\top \mb{K} - \mb{I}_{0}\|^2_F,
\label{eq:orth_loss}
\end{equation}
where $L_{\text{task}}$ is is the task loss and $\|\mb{K}^\top \mb{K} - \mb{I}_{0}\|^2_F$ is the orthogonality constraint (Eq.~\eqref{eq:orth_loss_row}) based on these DBT kernels, and $D$ is the total number of convolutional layers. This formulation not only addresses the traditional constraints but also considers the unique interactions of DBT-based orthonormality across different network blocks.

\subsection{Structured Orthogonal Regularization Optimization}
\label{sec:soro}

In Sec.~\ref{sec:3_2}, we discuss that the condition number of $\mb{G}$ ($\kappa\geq1$) in the linear system $\mb{g}=\mb{G}\mb{W}$ influences the training stability of the gradient descent process in multi-task learning. 
The condition number, a measure of sensitivity to changes in input, directly affects training stability. Specifically, a high condition number in the gradient matrix $\mb{G}$ indicates potential instability due to disproportionate scaling of gradient contributions during optimization~\cite{senushkin2023independent}.
The condition number is defined as:
\newtheorem{domin}{Definition}
\begin{domin}[Condition number of gradient system]
\label{label}
The condition number of the gradient matrix, formed from the gradients of the shared parameters:
\begin{equation}
    \kappa(\mb{G}) = \frac{\sigma_{\mathrm{max}} (\mb{G}_{\theta})}{\sigma_{\mathrm{min}} (\mb{G}_{\mb{\theta}})},
    \quad \mathrm{where  }~~  \mb{G}_{\theta} = [\mb{g}_{\mathrm{new}, 1}, \cdots, \mb{g}_{\mathrm{new}, n}, \mb{g}_{\mathrm{old}, 1}, \cdots, \mb{g}_{\mathrm{old}, m}].
\end{equation}
\end{domin}
The condition number describes how the relative scale of the largest and smallest singular values (represented by $\sigma_{\text{max}}$ and $\sigma_{\text{min}}$, respectively) can influence the effectiveness of gradient updates. 
When the condition number of the gradients is constrained to 1~\cite{senushkin2023independent}, we have
\begin{equation}
    \|\mb{g}_i\| \cdot \|\mb{g}_j\| = 0 \quad \text{and} \quad \|\mb{g}_i\| \cdot \|\mb{g}_i\| = \|\mb{g}_j\| \cdot \|\mb{g}_j\| = \mb{I}.
\label{eq:orth_grad_2}
\end{equation}
This means any two gradients are orthogonal.

In PCL scenario, differing gradients across tasks may cause dominance and conflict issues. Constraining the condition number of the gradient system $\mb{G} = [\mb{g}_{\text{new}, 1}, \cdots, \mb{g}_{\text{new}, n}, \mb{g}_{\text{old}, 1}, \cdots, \mb{g}_{\text{old}, m}]$ to 1 risks normalizing all gradients to the same scale, which may stifle learning by reducing natural variations. Our approach maintains mutual orthogonality and a low condition number while preserving the diversity of gradient directions. This strategy ensures stability and accommodates the unique contributions of each task, effectively reducing conflicts and dominance within the gradient system.

\begin{figure}[t]
\begin{minipage}{\linewidth}
\begin{algorithm}[H]
\caption{{Adjust gradient matrix(AdjustGradient)}.}
\label{alg:comp_g}
\LinesNumbered
\KwIn{Gradient matrix $\bm{G}$; $\text{diag}(\mb{d})$; Number of iterations $\mb{N}$; Step size: $\eta$}
\KwOut{$\bm{\hat{G}}$}

$\mb{U}, \mb{V} \leftarrow \text{SVD}(\mb{G})$\;
Initialize $\mb{d}_0 = \mb{1}$\ and $\mb{\hat{G}}_0 = \mb{U}\mb{V}^{T}$\;
\For{$k \in \mb{N}$}{
$\mb{\hat{G}}_{k} \leftarrow \mb{\hat{G}}_{k-1} -\eta \nabla_{\mb{\hat{G}}_{k-1}} ( \| \mb{G} - \mb{\hat{G}}_{k-1} \|^2_F + \sigma \| \mb{\hat{G}}_{k-1}^T \mb{\hat{G}}_{k-1} - \text{diag}(\mb{d}_{k-1}) \|^2_F)$\;
\tcp{Update $\mb{\hat{G}}_{k}$ by applying gradient descent}
$\mb{d}_{k} \leftarrow \frac{1}{\lambda + \sigma} (\lambda \mathbf{1} + \sigma \text{diag}(\mb{\hat{G}}_{k}^T \mb{\hat{G}}_{k}))\;
$
\tcp{Update $\mb{d}_{k}$ based on current $\mb{\hat{G}}_{k}$.}
}
\end{algorithm}
\end{minipage}
\end{figure}

Thus, we focus on the following optimization problem, where $\lambda$ is the hyperparameter constraining the condition number of $\hat{\mb{G}}$:
\begin{equation}
    \min\limits_{\hat{\mb{G}}, \kappa(\hat{\mb{G}})} \| \mb{G}-\hat{\mb{G}} \|^2_F + \lambda\kappa(\hat{\mb{G}}).
\label{eq:orgin}
\end{equation}
The scenario in Eq.~\eqref{eq:orth_grad_2} exemplifies this optimization target, where $\kappa(\mb{\hat{G}})=1$ means achieving the minimal condition number for the gradient system and indicative of an optimal state. This situation corresponds to a Procrustes problem~\cite{schonemann1966generalized}, offering a closed-form solution discussed in detail in Appendix~\ref{sec:opt_pro}. The solution can be derived using Singular Value Decomposition (SVD):
\begin{equation}
    \hat{\mb{G}} = \mb{U} \mb{V}^\top,  \quad \text{where~~} \mb{U}, \mb{V} : \mb{G}=\mb{U} \mb{\Sigma} \mb{V}^\top.
\label{eq:Procrustes}
\end{equation}
Given the challenges of strictly orthogonalizing gradients, we propose focusing on optimizing Eq.~\eqref{eq:orgin}. However, this non-convex problem is difficult to solve directly due to the high dimensionality of $\mb{G}$ and the computational demands in deep learning. Moreover, in PCL, strictly enforcing gradient orthogonality can restrict the model’s representational capacity~\cite{huang2018orthogonal}.
We have revised the optimization to Eq.~\eqref{eq:block_descent} where we no longer require strict orthogonality. By employing a large penalty term $\sigma$ to $\| \mb{\hat{G}}^\top \mb{\hat{G}} - \text{diag}(\mb{d}) \|^2_F$, we ensure $\hat{\mb{G}}^\top \hat{\mb{G}}$ is close to a diagonal matrix, achieving approximate orthogonality among gradients without mandating equal norms. Additionally, the second term $\| \mb{d} - \mb{1} \|^2_2$ constraints $\text{diag}\mb{d}$ to a diagonal matrix, with the regularization coefficient $\lambda$ allowing variations in the value of $\text{diag}(\mb{d})$.
The situation discussed in Eq.~\eqref{eq:Procrustes} can be considered as a special case when both penalty coefficient $\sigma$ and regularization coefficient $\lambda$ are set very large.
\begin{equation}
    \min\limits_{\mb{\hat{G}}, \mb{d}} \| \mb{G}- \mb{\hat{G}} \|^2_F + \sigma \| \mb{\hat{G}}^\top \mb{\hat{G}} - \text{diag}(\mb{d}) \|^2_F + \lambda \| \mb{d} - \mb{1} \|^2_2.
\label{eq:block_descent}
\end{equation}
Consequently, we transform the problem into a convex optimization challenge involving the variables $\hat{\mb{G}}$ and $\text{diag}(\mb{d})$, which can ideally be solved using the Block Coordinate Descent (BCD) method~\cite{chong2013introduction}. However, while descending on $\hat{\mb{G}}$, we are unable to obtain a closed-form solution, and embedding further optimization within this step significantly hinders computational efficiency. From an engineering perspective and inspired by~\cite{yi2016fast},we thus employ an alternating gradient descent approach to find solutions. Given that we can anticipate an excellent initial point, specifically $\mb{\hat{G}}_0 = \mb{U}\mb{V}^\top$ and $\text{diag}(\mb{d}) = \mb{I}$, we utilize Alg.~\ref{alg:comp_g} to achieve convergence to a solution with a decrease below $10^{-3}$ per iteration within an acceptable computation time. The penalty parameter $\sigma$ is critical for ensuring the orthogonality and proper scaling of the columns of $\mb{\hat{G}}$, as discussed in Appendix~\ref{sec:hyper}.

\subsection{Algorithm}

\begin{wrapfigure}[16]{r}{0.55\linewidth}
\vspace{-30px}
\centering
\begin{minipage}{.95\linewidth}
\begin{algorithm}[H]
\caption{{Stable Parallel Continual Learning}.}
\label{alg:SPCL}
\LinesNumbered
\KwIn{
Parameters $\bm{\theta}$, $\bm{\theta}_{1:T}$; Step sizes $\gamma$, $\gamma_{1:T}$, $\eta$ 
}
\KwOut{$\bm{\theta}$, $\bm{\theta}_{1:T}$}
\For{$t$ in timeline}{
$\mc{T}_t\leftarrow$ in-training task index\;
\For{$i\in\mc{T}_t$}{
$\mc{B}_i\sim\mc{D}_i$\;
$\nabla_{\bm{\theta_i}} \ell_i\left(\mc{B}_i\right) = \nabla_{\bm{\theta_i}} \ell_i\left(\mc{B}_i\right)_{\text{task}} + \alpha \nabla_{\bm{\theta_i}} \ell_i\left(\mc{B}_i\right)_{\text{orth}}$\;
$\bm{\theta}_i\leftarrow\bm{\theta}_i-\gamma_i \nabla_{\bm{\theta_i}}\ell_i\left(\mc{B}_i\right) $\;
$\mb{g}_i=-\nabla_{\bm{\theta}}\ell\left(\mc{B}_i\right)$\;
}
$\mb{\hat{G}}_t\leftarrow \text{AdjustGradient}([\mb{g}_i]), \forall_i$\;
$\bm{\theta}\leftarrow\bm{\theta}+\gamma\cdot\mb{\hat{G}}_t \mb{W}$\;\tcp{W = I}
}
\end{algorithm}
\end{minipage}
\end{wrapfigure}

The SPCL approach outlined in Alg.~\ref{alg:SPCL} utilizes advanced gradient and kernel orthogonalization methods to improve learning stability in neural networks.
For each task and at each timestep, the network computes gradients for both task-specific loss and orthogonal regularization loss. 
These gradients are then adjusted for orthogonality using the method from Sec.~\ref{sec:soro} before updating the model parameters. 
In the process of computing $\mb{\hat{G}}$ in Alg.~\ref{alg:SPCL}, the original constrained optimization problem Eq.~\eqref{eq:orgin}, is relaxed into an unconstrained optimization problem Eq.~\eqref{eq:block_descent} using the penalty function method. This method introduces a penalty term $\|\mb{\hat{G}}^T\mb{\hat{G}} - \text{diag}(\mb{d})\|^2_F$ to the objective function to handle constraints effectively.

\begin{table*}[t]
\centering
\caption{
Task-incremental PCL final performance comparisons (avg $\pm$ std). 
}
\resizebox{\linewidth}{!}{
\begin{tabular}{l|c|rr|rr|rr}
\toprule
\textbf{Method} & \multirow{2}{*}{\textbf{Type}} &  \multicolumn{2}{c|}{\textbf{PS-EMNIST}} &  \multicolumn{2}{c|}{\textbf{PS-CIFAR-100}} &  \multicolumn{2}{c}{\textbf{PS-ImageNet-TINY}} \\
(Rehearsal) & & \multicolumn{1}{c}{$A_{\bar{e}}$ (\%)} & \multicolumn{1}{c|}{$F_{\bar{e}}$ (\%)} & \multicolumn{1}{c}{$A_{\bar{e}}$ (\%)} & \multicolumn{1}{c|}{$F_{\bar{e}}$ (\%)}& \multicolumn{1}{c}{$A_{\bar{e}}$ (\%)} & \multicolumn{1}{c}{$F_{\bar{e}}$ (\%)}  \\
\midrule
MGDA~\cite{desideri2012multiple} & MTL & $84.887\pm0.469$ & $-4.301\pm0.818$ & $48.957\pm0.451$ & $25.088\pm0.203$  & $38.058\pm0.740$ & $1.465\pm0.490$ \\
DWA~\cite{liu2019end}   & MTL      & $88.405\pm0.322$ & $-5.452\pm0.338$ & $44.969\pm0.378$ & $22.682\pm0.537$ & $34.290\pm1.099$ & $-1.053\pm0.866$ \\
PCGrad~\cite{yu2020gradient} & MTL  & $89.698\pm0.164$ & $-4.921\pm0.121$ &  $47.026\pm0.538$ & $23.244\pm0.740$ & $39.427\pm1.275$ & $2.017\pm0.769$ \\
RLW~\cite{lin2021closer} 	      & MTL      & $89.288\pm0.218$ & $-5.226\pm0.204$ & ${47.574}\pm0.349$ & $23.833\pm0.117$ & $38.531\pm1.610$ & $1.332\pm1.148$ \\
\midrule
AGEM~\cite{chaudhry2018efficient} & SCL	 & $87.022\pm0.519$ & $-7.646\pm0.483$  & $27.379\pm0.585$ & $5.416\pm0.851$ & $28.530\pm0.994$ & $-7.070\pm1.410$ \\
GMED~\cite{jin2021gradient} & SCL	 & $85.471\pm0.324$ & $-8.875\pm0.335$  & $49.094\pm1.792$&$18.356\pm1.345$& $34.495\pm1.568$ & $-0.640\pm1.799$  \\
ER~\cite{chaudhry2019on} & SCL	 & $89.106\pm0.315$ & $-5.525\pm0.207$ & $47.324\pm0.584$ & $23.330\pm0.762$ & $35.950\pm0.763$ & $-0.767\pm0.591$ \\
MEGA~\cite{guo2019learning} & SCL	 & $86.262\pm0.582$ & $-4.753\pm0.295$ & ${47.745}\pm0.460$ & $23.718\pm0.832$ & $38.386\pm1.234$ & $2.655\pm0.824$\\
MDMTR~\cite{lyu2021multi} & SCL	 & $73.115\pm0.382$ & $-16.359\pm0.178$ & $38.865\pm1.691$  & $18.297\pm1.052$ &$23.422\pm0.273$ & $-1.544\pm0.389$\\
DER++~\cite{buzzega2020dark} & SCL	 & $89.654\pm0.451$ & $-4.979\pm0.293$  &$49.276\pm0.480$&$22.634\pm0.165$& $40.039\pm0.039$ & $2.601\pm0.643$\\
\midrule
MaxDO~\cite{lyu2023measuring} & PCL	 & $90.189\pm0.314$ & $-4.258\pm0.311$  &$50.203\pm0.978$&$24.510\pm0.092$& $40.770\pm0.354$ & $3.119\pm0.450$\\
SORO & PCL & $91.062\pm0.447$ & $-4.679\pm0.425$ & $54.728\pm0.687$ & $27.139\pm0.137$ & $47.445\pm0.645$ & $7.948\pm0.704$\\
SORO+DBT$_{orth}$ & PCL & $91.425\pm0.241$ & $-3.908\pm0.093$ & $54.788\pm0.805$ & $27.476\pm0.644$ & $47.961\pm0.062$ & $7.734\pm0.305$\\
\bottomrule
\end{tabular}}
\label{tab:main_task}
\end{table*}

\begin{table*}[t]
\centering
\caption{
Class-incremental PCL final performance comparisons (avg $\pm$ std). 
}
\resizebox{\linewidth}{!}{
\begin{tabular}{lcrrrrrr}
\toprule
\textbf{Method} & \multirow{2}{*}{\textbf{Type}} &  \multicolumn{2}{c}{\textbf{PS-EMNIST}} &  \multicolumn{2}{c}{\textbf{PS-CIFAR-100}} &  \multicolumn{2}{c}{\textbf{PS-ImageNet-TINY}} \\
(Rehearsal) & & \multicolumn{1}{c}{$A_{\bar{e}}$ (\%)} & \multicolumn{1}{c}{$F_{\bar{e}}$ (\%)} & \multicolumn{1}{c}{$A_{\bar{e}}$ (\%)} & \multicolumn{1}{c}{$F_{\bar{e}}$ (\%)}& \multicolumn{1}{c}{$A_{\bar{e}}$ (\%)} & \multicolumn{1}{c}{$F_{\bar{e}}$ (\%)}  \\
\midrule
MGDA~\cite{desideri2012multiple} & MTL & $52.823\pm0.201$ & $-7.753\pm0.661$  & $11.323\pm0.282$ & $-4.065\pm0.725$ & $8.318\pm0.155$ & $-13.118\pm0.739$\\
DWA~\cite{liu2019end}   & MTL      & $42.734\pm0.211$ & $-38.944\pm0.581$ & $3.519\pm0.116$ & $-7.470\pm0.244$ & $3.003\pm0.139$ & $-12.774\pm0.158$\\
PCGrad~\cite{yu2020gradient} & MTL  & $45.035\pm0.273$ & $-43.309\pm0.176$ & $10.140\pm0.253$ & $-5.149\pm0.553$ & $7.789\pm0.237$ & $-15.860\pm0.318$\\
RLW~\cite{lin2021closer} 	      & MTL      & ${44.595}\pm0.480$ & $-43.424\pm0.325$ & $9.957\pm0.135$ & $-5.173\pm0.613$ &$7.419\pm0.258$&$-16.079\pm0.255$ \\
\midrule
AGEM~\cite{chaudhry2018efficient} & SCL	 & $26.249\pm0.511$ & $-62.480\pm0.527$ & $3.374\pm0.098$ & $-10.462\pm0.543$ & $3.379\pm0.149$ & $-18.514\pm0.516$\\
GMED~\cite{jin2021gradient} & SCL	 & $22.694\pm0.153$&$-65.805\pm0.255$& $4.941\pm0.407$ & $-12.837\pm0.710$  & $3.386\pm0.348$ & $-19.060\pm0.682$ \\
ER~\cite{chaudhry2019on} & SCL	  & $43.474\pm0.860$ & $-44.597\pm0.861$ & $9.317\pm0.339$ & $-5.337\pm0.626$ & $6.126\pm0.277$ & $-13.686\pm0.290$\\
MEGA~\cite{guo2019learning} & SCL	 & ${51.797}\pm0.607$ & $-5.358\pm0.865$ & $12.245\pm0.388$ & $0.380\pm0.503$ & $10.082\pm0.302$ & $-2.375\pm0.119$ \\
MDMTR~\cite{lyu2021multi} & SCL	  & $42.199\pm1.100$  & $-32.140\pm0.996$ &$5.525\pm0.260$ & $-3.454\pm0.145$ & $2.793\pm0.042$ & $-12.992\pm0.082$\\
DER++~\cite{buzzega2020dark} & SCL	 & $50.039\pm0.520$ & $-43.422\pm0.439$  &$11.022\pm0.642$&$-1.653\pm0.120$&$9.190\pm0.429$&$-14.049\pm0.227$\\
\midrule
MaxDO~\cite{lyu2023measuring} & PCL	 & $53.139\pm0.156$ & $-11.903\pm0.476$  &$12.237\pm0.176$&$-2.280\pm0.270$& $9.532\pm0.363$ & $-12.511\pm0.610$\\
SORO & PCL & $53.508\pm0.416$ & $-30.788\pm0.240$ & $15.081\pm0.950$ & $0.222\pm0.221$ & $12.056\pm0.474$ & $-12.605\pm0.375$\\
SORO+DBT$_{orth}$ & PCL & $53.738\pm0.118$ & $-31.836\pm0.669$ & $15.870\pm0.132$ & $3.010\pm0.131$ & $12.470\pm0.058$ & $-12.170\pm0.575$\\
\bottomrule
\end{tabular}}
\label{tab:main_class}
\end{table*}

\section{Experiment} 
\label{sec:exp_detl}

\subsection{Dataset}

Following~\cite{lyu2023measuring}, we evaluate on three PCL datasets in both task-incremental and class-incremental learning scenarios.
\\
\textbf{Parallel Split EMNIST} (PS-EMNIST), which includes 62 classes and is divided into 5 tasks. Each task is assigned three random label sets, with three timelines per label set, resulting in 9 different configs, which is same as the other two datasets. \\
\textbf{Parallel Split CIFAR-100} (PS-CIFAR-100), which includes 100 classes, is divided into 20 tasks.\\ 
\textbf{Parallel Split ImageNet-TINY} (PS-ImageNet-TINY) includes 200 classes, is divided into 20 tasks.

\subsection{Experiment details}

We set the batch size uniformly at 128 across all datasets, employing a 2-layer MLP for PS-EMNIST and a Resnet-18~\cite{he2016deep} for PS-CIFAR-100 and PS-ImageNet-Tiny. Our rehearsal strategy maintains varying samples per class: 30 for PS-CIFAR-100, 5 for PS-EMNIST, and 15 for PS-ImageNet-Tiny.
For evaluating PCL, we measure average accuracy and forgetting, following methodologies from previous studies such as~\cite{lopez2017gradient, chaudhry2018efficient}. With $\bar{e} = \max (e_1, e_2, \cdots, e_T)$, we calculate:
\begin{equation}
    A_{\bar{e}} = \frac{1}{T} \sum\nolimits_{t=1}^{T} a_{\bar{e}}^{t}\quad F_{\bar{e}} = \frac{1}{T} \sum\nolimits_{t=1}^{T} a_{\bar{e}}^{t} - a_{e_{t}}^{t}
\end{equation}
Here, $A_{\bar{e}}$ signifies the overall average accuracy across tasks, while $F_{\bar{e}}$, or backward transfer, indicates the drop in performance relative to the initial training of each task.
All models are deployed with tensorflow and the experiments are conducted on an RTX 4090 GPU.

\subsection{Main Results}

In our study, we benchmark SPCL against various MTL approaches such as MGDA~\cite{sener2018multi}, DWA~\cite{liu2019end}, GradDrop~\cite{chen2020just}, PCGrad~\cite{yu2020gradient}, and RLW~\cite{lin2021closer} within a PCL framework. We treat each moment as a distinct MTL segment to facilitate PCL training, implementing each MTL method according to specifications from their respective publications. Additionally, we draw comparisons with rehearsal-based SCL methods like AGEM~\cite{chaudhry2018efficient}, GMED~\cite{jin2021gradient}, and ER~\cite{chaudhry2019tiny}. To align with SCL strategies, we combine batches from all active tasks, simulating a straightforward sequential learning setup where only batches from the current task and a memory buffer are used. This methodological adaptation allows us to assess the efficacy of SCL methods in a PCL context effectively.
We show the main comparisons with the proposed methods in Tables~\ref{tab:main_task} and~\ref{tab:main_class}  on the three datasets. We have several major observations.
Our methods shows significant improvements over others in both task-incremental and class-incremental PCL scenarios, especially on the PS-CIFAR-100 and PS-ImageNet-TINY dataset.
Compared to Structured Orthogonal Regularization Optimization(SORO), which only orthogonalizes gradients during PCL model training, incorporating orthogonality constraints on CNN filters further enhances model performance. The combination of these two methods provides more stable training.

\subsection{SPCL learning process}

\begin{figure}[t]
\centering
\subfigure[PS-CIFAR-100]{\centering
  \includegraphics[width=0.48\textwidth]{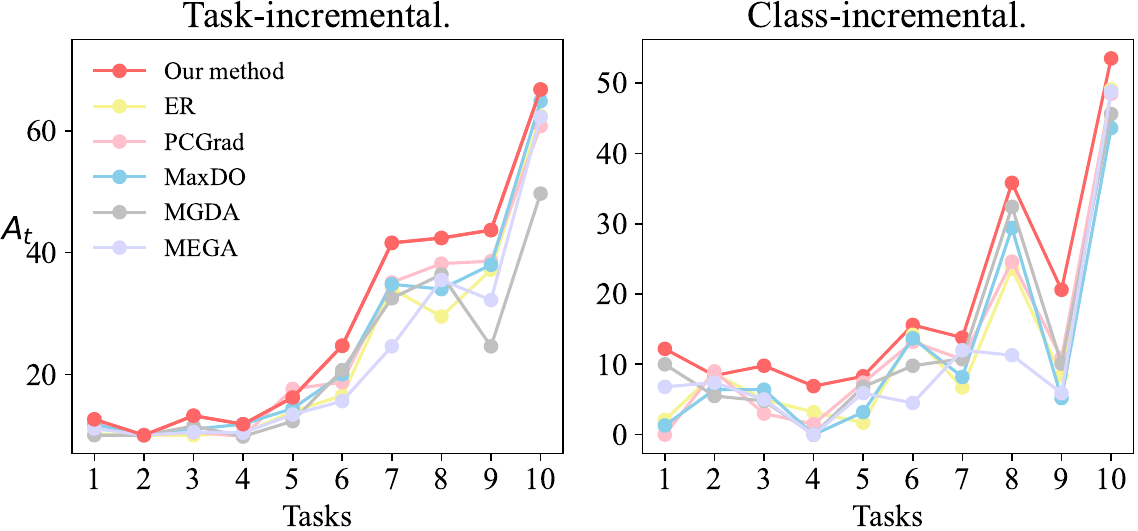}}
\subfigure[PS-ImageNet-TINY]{\centering
  \includegraphics[width=0.48\textwidth]{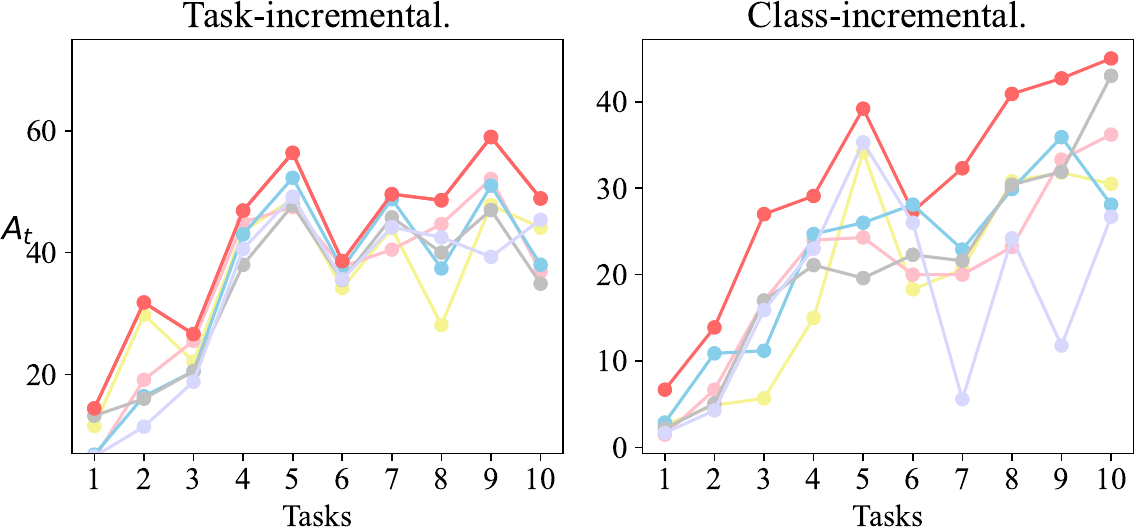}}
\caption{Learning process comparisons on two datasets.}
\label{fig:acc_line_}
\end{figure}

\begin{figure}[h]
  \centering
  \includegraphics[width=0.38\textwidth]{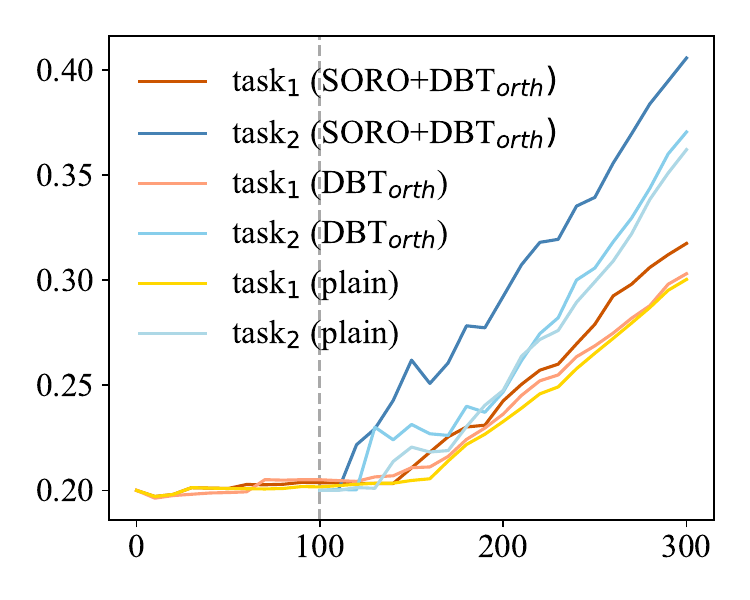} 
  \caption{Ablation study on CIFAR-10: This involves two tasks, with the dashed line indicating the start of the second task, depicting the progression of accuracy over time for two tasks under different orthogonalization conditions.}
\label{fig:acc_line_comp}
\end{figure}

Thge ablation study on the CIFAR-10 dataset is shown in Fig.~\ref{fig:acc_line_comp}, which validates the efficacy of orthogonal regularization in managing task interference in PCL models. 
By applying orthogonal constraints either on the gradients , the filter in CNN, or both, the model shows varied capability in preserving learning stability and improving task-specific accuracy over training epochs. The introduction of Task 2 is a critical moment, as indicated by the dashed line, where the model's ability to adapt to new tasks while retaining performance on ongoing tasks is evaluated. 
The results show the orthogonalization enhances the adaptiveness and stability of continual learning systems.

Fig.~\ref{fig:acc_line_} tracks the performance of seen tasks throughout the PCL training process. Each point represents the completion of training for the task, showing the average testing accuracy for all ongoing and completed tasks. Our SPCL significantly improved performance across three datasets in both task-incremental and class-incremental PCL settings, consistently outperforming baseline models, demonstrating its robustness and generalizability.

\begin{figure}[t]
\centering
\subfigure[Condition number]{\centering
  \includegraphics[width=0.325\textwidth]{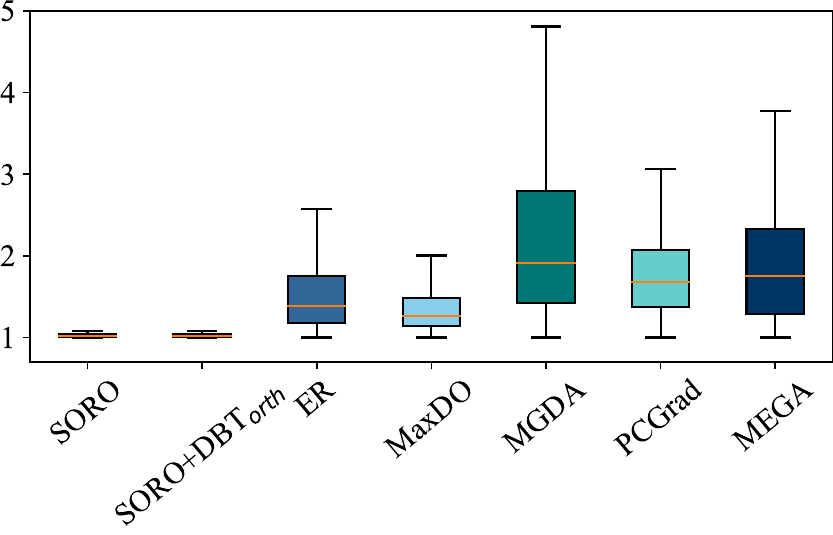}}
\subfigure[Gradient conflicts]{\centering
  \includegraphics[width=0.325\textwidth]{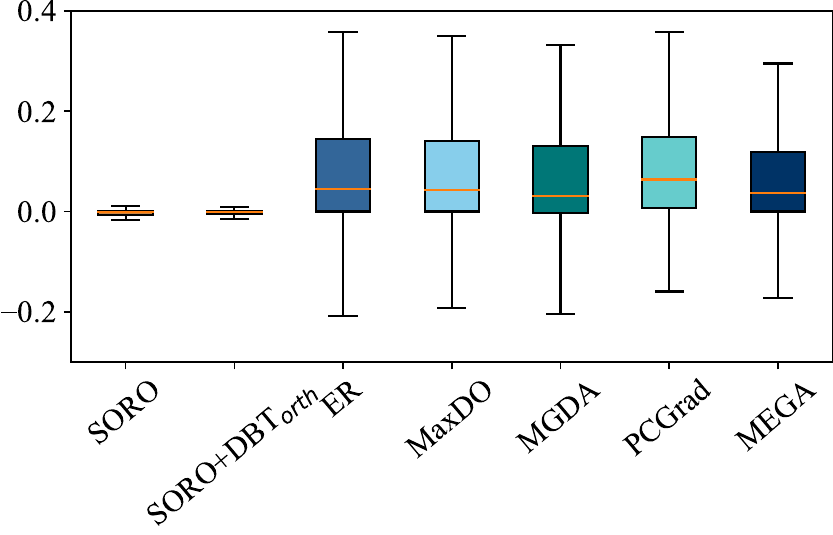}}
\subfigure[Gradient magnitude similarity]{\centering
  \includegraphics[width=0.325\textwidth]{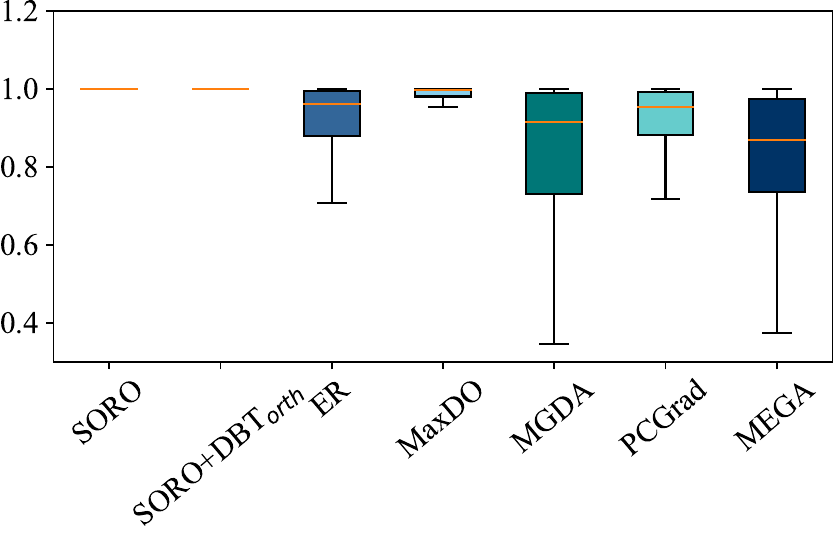}}
\caption{Numerical analysis of gradient matrix on PS-CIFAR-100.}
\label{fig:numerical_grad}
\end{figure}

We empirically analyze condition number,cosine distance and gradient magnitude similarity~\cite{yu2020gradient} using PCL on PS-CIFAR-100. The gradient magnitude similarity measure indicates that ER~\cite{chaudhry2019tiny}, MGDA~\cite{sener2018multi} and MEGA~\cite{guo2019learning} suffer from gradient dominance. Interestingly, MaxDO~\cite{lyu2023measuring}, despite performing well, is misleadingly flagged as problematic by the gradient cosine distance measure. In contrast, the condition number, as a stability criterion, highlights domination issues for MGDA~\cite{sener2018multi} and MEGA~\cite{guo2019learning}. As shown in Fig.~\ref{fig:numerical_grad}, the condition number more clearly exposes training issues, and the condition number of the gradients obtained by our method is very close to 1.

\subsection{Numerical analysis}
\label{sec:exp_num_any}

\begin{figure}[t]
  \centering
  \includegraphics[width=0.48\textwidth]{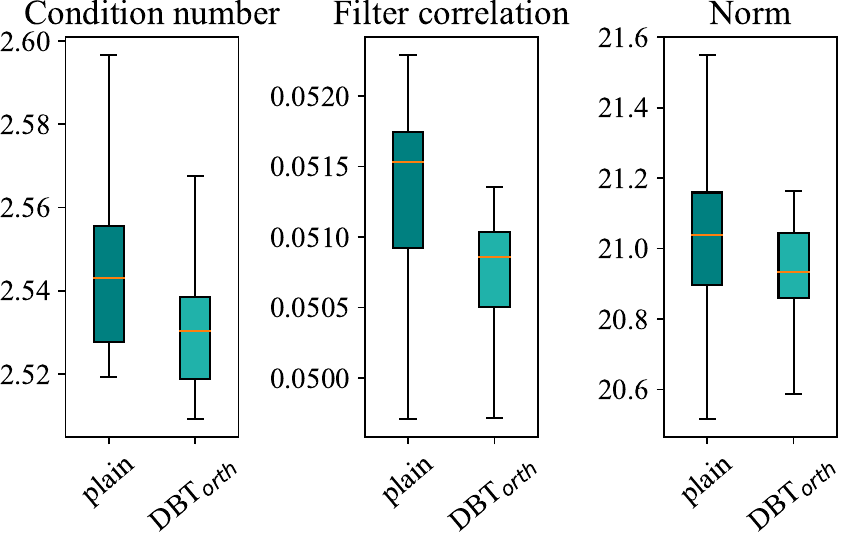} 
  \caption{Numerical analysis of CNN convolutional layers after adopting a DBT-based orthogonality regularization method on cifar-100.}
\label{fig:numerical_cnn}
\end{figure}

We also compare the effects of introducing orthogonality constraints via a penalty function in the convolutional layers of CNNs. We contrast the condition number distribution, norm distribution, and filter redundancy of convolutional layers with and without the DBT-based orthogonalization method. In Fig.~\ref{fig:numerical_cnn}, experimental results indicate that employing DBT-based orthogonalization improves the stability of convolutional layer parameters.
Feature redundancy presents a significant issue, where different convolutional kernels may learn highly similar features. Filter correlation is defined as:
\begin{equation}
\frac{1}{m(m-1)} \sum_{1\leq i < j \leq m} | \frac{\mb{W}_{i, :} \cdot \mb{W}_{j, :}}{\|\mb{W}_{i, :} \| \|\mb{W}_{j, :} \|} |
\end{equation}
Our experiments demonstrate that by applying orthogonality constraints to the convolutional layers in the backbone CNN network, we can reduce the condition number of the parameter layer matrices, decrease filter correlation between parameters, and reduce the norms of the parameters.

\section{Conclusion}
\label{sec:conclusion}

In conclusion, this paper systematically investigates the benefits of orthogonalization in both forward and backward propagation within the context of PCL. Through our studies, we have identified that entangled features during forward propagation can result in conflicting activations, whereas gradient updates from concurrent tasks often lead to conflicts in backward propagation. To address these challenges, we introduce a dual-strategy approach named SPCL, which incorporates orthogonality at two stages of the learning process.
The first strategy in our approach applies DBT Matrix-based orthogonality constraints directly to the network's filters, ensuring stable and consistent forward propagation. The second strategy employs orthogonal decomposition to manage gradients, effectively stabilizing backward propagation and reducing conflicts among task-specific gradients. Both strategies utilize soft orthogonal regularization to optimize the balance between computational efficiency and robust training dynamics, enhancing the overall effectiveness of the learning process.
However, a \textit{limitation} of our approach is the need for precise tuning of the orthogonality regularization parameter, which is sensitive to task characteristics. Incorrect settings may lead to excessive or insufficient orthogonalization, impacting model adaptability and increasing computational time due to convergence difficulties. Future researches should explore adaptive methods for dynamically adjusting this parameter to improve model flexibility and efficacy in response to task complexities and data variations.

\bibliographystyle{unsrtnat}

\appendix

\section{Appendix}
\label{sec:Appendix}

\subsection{Overview of Learning Paradigms: MTL, SCL, and PCL}

\begin{figure}[h]
\centering
\subfigure[Multi-Task Learning]{\centering
  \includegraphics[width=0.325\textwidth]{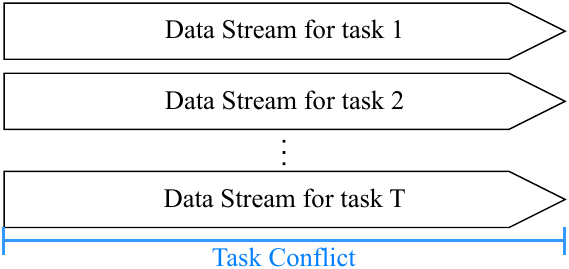}}
\subfigure[Serial Continual Learning]{\centering
  \includegraphics[width=0.325\textwidth]{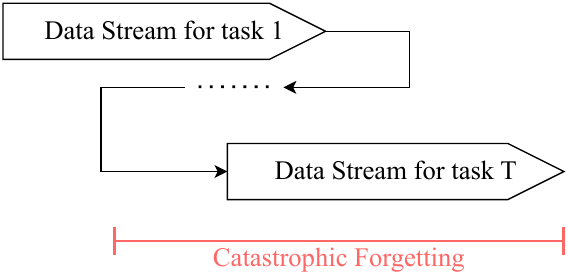}}
\subfigure[Parallel Continual Learning]{\centering
  \includegraphics[width=0.325\textwidth]{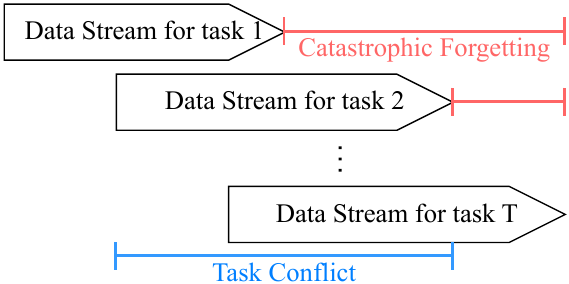}}
\caption{Comparisons of Multi-Task Learning, Serial Continual Learning and Parallel Continual Learning.}
\label{fig:compare_pcl}
\end{figure}

As shown in Fig. \ref{fig:compare_pcl}, Multi-Task Learning (MTL), Serial Continual Learning (SCL, or traditional CL), and Parallel Continual Learning (PCL) are three paradigms addressing the challenges of learning from multiple tasks. MTL simultaneously trains on multiple tasks to leverage shared knowledge but faces task conflicts. SCL sequentially trains tasks to avoid simultaneous complexity, but struggles with catastrophic forgetting, where new learning can erase previous knowledge. PCL combines these approaches, allowing for simultaneous training of new tasks as they arise, aiming to dynamically manage both task conflict and catastrophic forgetting, thus fitting more realistic and varied learning scenarios.

\subsection{Doubly-block Toeplitz Matrix}
\label{sec:DBTM}

The linear transformation between layers in CNN is computed as $\mb{Y}=\mb{K} \ast \mb{X}$. Alternatively, the kernel can be expanded using a faster doubly-block Toeplitz (DBT) matrix~\cite{wang2020orthogonal}.
DBT regularization enhances convolutional neural networks by structuring convolution as a DBT matrix-vector multiplication, which reduces storage requirements from $O(n^2)$ to $O(n)$ and computational complexity from $O(n^2)$ to $O(n\log n)$~\cite{osahor2022ortho}.

A Toeplitz matrix is characterized by constant elements along each descending diagonal from left to right. A DBT matrix extends this concept by being a blocked Toeplitz matrix where each block is itself a Toeplitz matrix. Let $vec(\mb{X})$ denote as flatten $\mb{X}$:
\begin{equation}
vec(\mb{X}) = [\mb{X}_{0}; \cdots; \mb{X}_{H-1}]^\top
\end{equation}
then, the convolution operation can be rewritten as:
\begin{equation}
vec(\mb{Y})=\mb{M}\cdot vec(\mb{X}) 
\end{equation}
where $\mb{M}$ is a DBT matrix is formatted as $\mb{R}^{O\times C\times H^{\prime}W^{\prime}\times HW}$:
\begin{equation}
\mb{M} = \begin{bmatrix}
    \mb{M}_{1, 1} & \mb{M}_{1, 2} & \cdots & \mb{M}_{1, C} \\
    \mb{M}_{2, 1} & \mb{M}_{2, 2} & \cdots & \mb{M}_{2, C} \\
    \vdots & \ddots & \vdots \\
    \mb{M}_{N, 1} & \mb{M}_{N, 2} & \cdots & \mb{M}_{N, C} \\
    \end{bmatrix}
\end{equation}
where each $\mb{M}_{i, j} \in R^{H^{\prime}W^{\prime}\times HW}$ is the i-th row and j-th column block matrix and we denote $Tope(\mb{w}_{n})$ as a Toeplitz matrix generated by n-th row vector of convolution kernel $\mb{w}$:
\begin{equation}
\mb{M}_{i, j} = \begin{bmatrix}
            Tope(\mb{w}_{0}) & \cdots & Tope(\mb{w}_{k-1}) \\
            & Tope(\mb{w}_{0}) & \cdots & Tope(\mb{w}_{k-1}) \\
            & & \cdots & \cdots \\
            & & & Tope(\mb{w}_{0}) & \cdots & Tope(\mb{w}_{k-1}) \\
            \end{bmatrix}
\end{equation}

\subsection{Preservation of Orthogonality in Linear Transformations.}

The preservation of orthogonality in linear transformations is first discussed in~\cite{huang2020controllable}:
\begin{theorem} \text{(Preservation of Orthogonality in Linear Transformations.)}
Let $\mb{\hat{h}} = \mb{W}\mb{x}$, where $\mb{W}^\top \mb{W} = \mb{I}$ and $\mb{W} \in \mathbb{R}^{n \times d}$.
Assume: (1) $\mathbb{E}_{\mb{x}}(\mb{x}) = 0$, $cov(\mb{x}) = \sigma_1^2 \mb{I}$, and (2) $\mathbb{E}_{\frac{\partial{\mathcal{L}}}{\partial{\mb{\hat{h}}}}}(\frac{\partial{\mathcal{L}}}{\partial{\mb{\hat{h}}}}) = \mb{0}$, $cov(\frac{\partial{\mathcal{L}}}{\partial{\mb{\hat{h}}}}) = \sigma_2^2 \mb{I}$. 
If $n=d$, we have the following properties: 
(1) $\| \mb{\hat{h}} \| = \| \mb{x} \|$; 
(2) $\mathbb{E}_{\mb{\hat{h}}}(\mb{\hat{h}}) = \mb{0}$, $cov(\mb{\hat{h}}) = \sigma_1^2 \mb{I}$;
(3) $\| \frac{\partial{\mathcal{L}}}{\partial{\mb{x}}} \| = \| \frac{\partial{\mathcal{L}}}{\partial{\mb{\hat{h}}}} \|$;
(4) $\mathbb{E}_{\frac{\partial{\mathcal{L}}}{\partial{\mb{x}}}}(\frac{\partial{\mathcal{L}}}{\partial{\mb{x}}}) = \mb{0}$, $cov(\frac{\partial{\mathcal{L}}}{\partial{\mb{x}}}) = \sigma_2^2 \mb{I}$.
In particular, if $n < d$, property (2) and (3) hold; if $n > d$, property (1) and (4) hold.
\label{thm:orth_2}
\end{theorem}

\subsection{Optimization Goal}
\label{sec:opt_pro}

\begin{figure}[t]
\centering
\subfigure{\centering
  \includegraphics[width=0.9\textwidth]{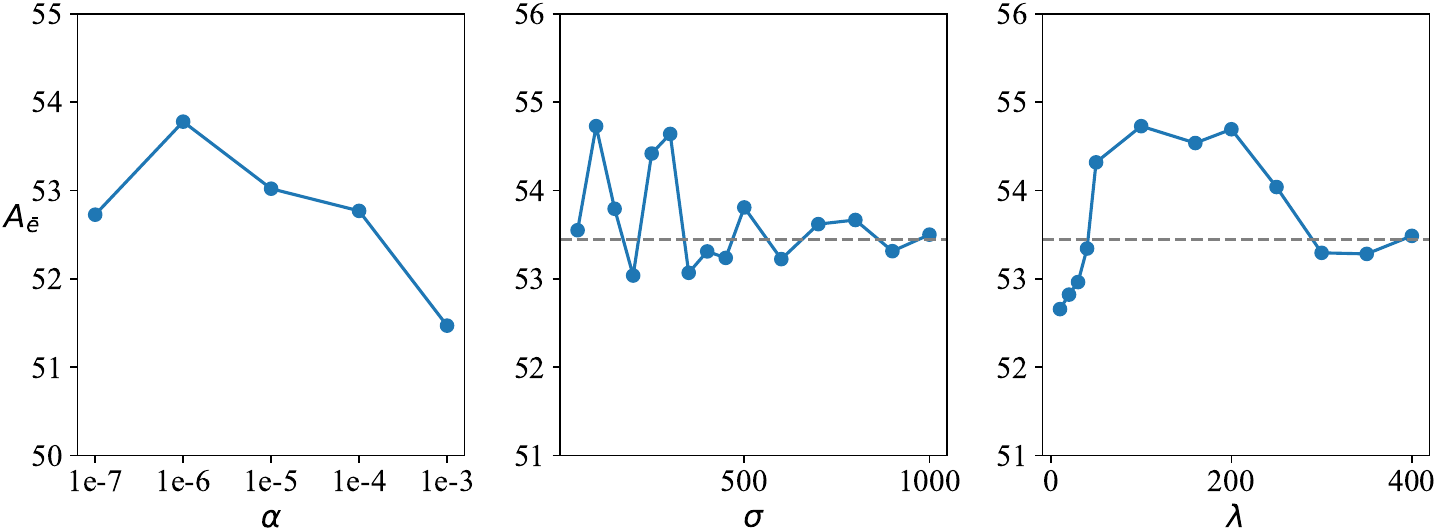}}
\caption{Hyperparameter selection.}
\label{fig:hparms}
\end{figure}
\begin{figure}[t]
\centering
\subfigure{\centering
  \includegraphics[width=0.5\textwidth]{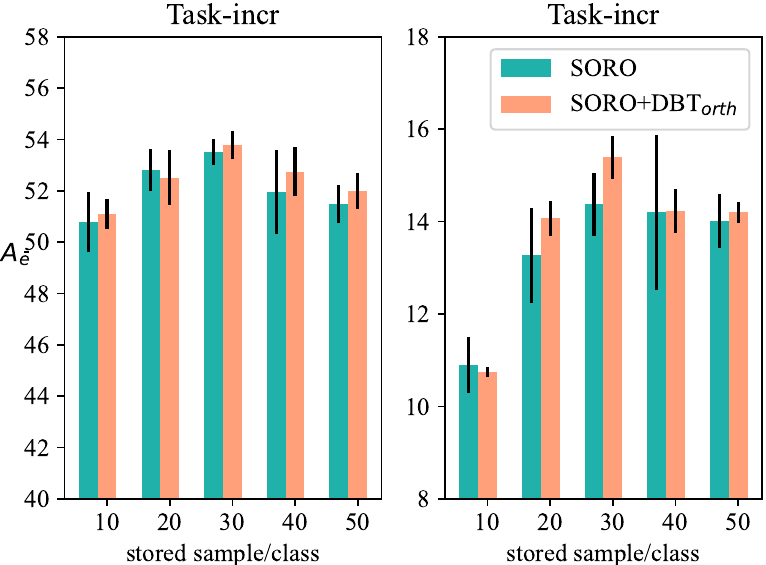}}
\caption{Reherasal analysis between SORO and $\text{SORO} + \text{DBT}_{orth}$ method.}
\label{fig:mem}
\end{figure}

Given a matrix $\mb{G} \in \mb{R}^{n\times t}$ representing the gradients of different tasks in a multi-task learning framework, where $\mb{t}$ is the number of tasks and $\mb{t} \ll \mb{n}$. We expect to solve the following optimization problem:
\begin{equation}
    \min\limits_{\hat{\mb{G}}} \| \mb{G}-\hat{\mb{G}} \|^2_F, \quad
    \text{s.t.} \hat{\mb{G}}^\top \hat{\mb{G}} = \mb{I}
\label{eq:procrus}
\end{equation}
this is a special kind of Procrustes problem~\cite{schonemann1966generalized}, so there exists a closed-form solution. To solve this problem, we first calculate the covariance matrix $\mb{M} = \mb{G}^{T}\mb{G}$ and perform SVD on $\mb{G} = \mb{U} \mb{\Sigma} \mb{V}^\top$. Therefor, we have:
\begin{equation}
\begin{aligned}
    \| \mb{G}-\hat{\mb{G}} \|^2_F 
    = &tr((\mb{G}-\hat{\mb{G}})^\top(\mb{G}-\hat{\mb{G}})) \\
    = &tr(\mb{G}^\top \mb{G} - \mb{G}^\top \hat{\mb{G}} - \hat{\mb{G}}^\top \mb{G} +  \hat{\mb{G}}^T \hat{\mb{G}}  ) \\
    = &tr(\mb{M}) + tr(\mb{I}) - tr(\hat{\mb{G}}^\top \mb{G}) - tr((\hat{\mb{G}}^\top \mb{G})^\top) \\
    = &tr(\mb{M}) + tr(\mb{I}) - 2tr(\hat{\mb{G}}^\top \mb{G})\\
    = &t + tr(\mb{M}) - 2tr(\hat{\mb{G}}^\top \mb{U} \mb{\Sigma} \mb{V}^\top) \\
    = &t + tr(\mb{M}) - 2tr(\hat{\mb{G}}^\top \mb{U} \mb{V}^\top \mb{\Sigma})
\label{eq:compute_g_}
\end{aligned}
\end{equation}
In the context of SVD, $\mb{\Sigma}$ is a diagonal matrix that contains the singular values $\mb{\sigma}_1, \mb{\sigma}_2, \cdots, \mb{\sigma}_t$ of the decomposed matrix, with non-diagonal entries being zero. In the equation $\hat{\mb{G}}^\top \mb{U} \mb{V}^\top$, each component is orthogonal, and their product will also result in an orthogonal matrix. Therefore, when $\hat{\mb{G}}^\top \mb{U} \mb{V}^\top  = \mb{I}$, where $\mb{I}$ is the identity matrix, the promblem in Eq.~\eqref{eq:procrus} reaches its optimal solution, indicating that $\mb{\hat{G}} = \mb{U}\mb{V}^T$. This represents the case where the transformation matrix $\mb{\hat{G}}$ is directly formed by the product of the matrices $\mb{U}$ and $\mb{V}$ from the SVD on $\mb{G}$.

\begin{figure}[t]
\centering
\subfigure{\centering
  \includegraphics[width=0.24\textwidth]{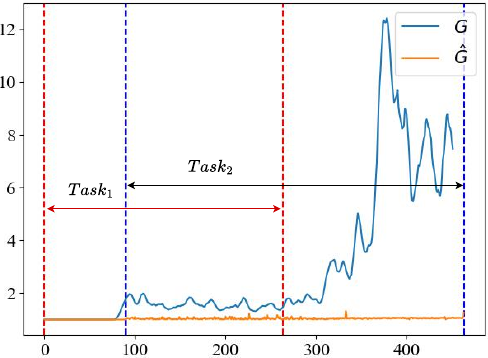}}
\subfigure{\centering
  \includegraphics[width=0.24\textwidth]{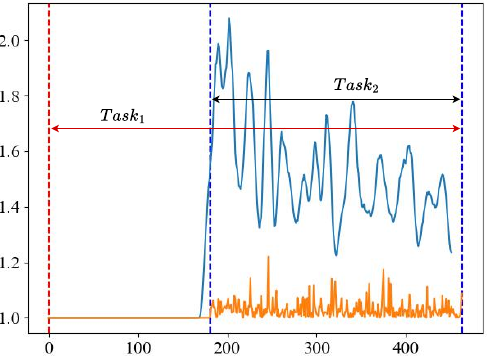}}
\subfigure{\centering
  \includegraphics[width=0.24\textwidth]{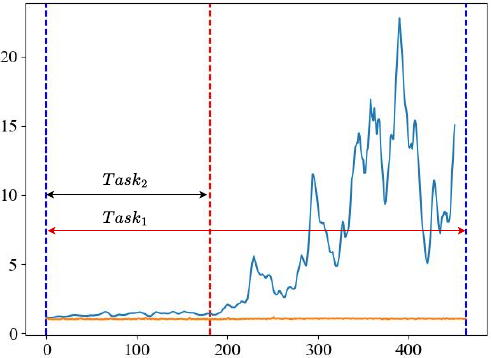}}
\subfigure{\centering
  \includegraphics[width=0.24\textwidth]{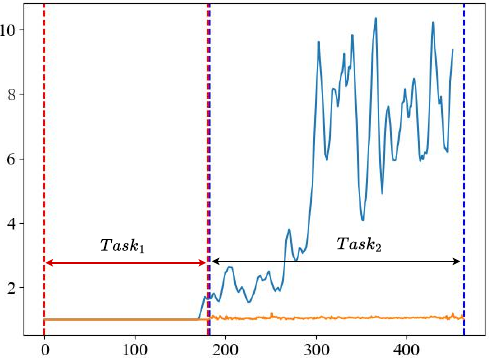}}
\caption{The variation in the condition number of the original gradient matrix $\mb{G}$ and adjusted gradient matrix $\hat{\mb{G}}$ under different timeline conditions.
}
\label{fig:conds_2}
\end{figure}

\begin{figure}[t]
\centering
\subfigure{\centering
  \includegraphics[width=0.24\textwidth]{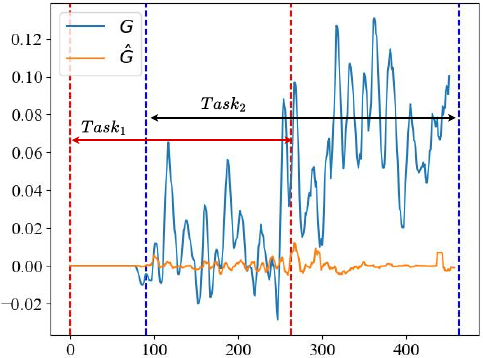}}
\subfigure{\centering
  \includegraphics[width=0.24\textwidth]{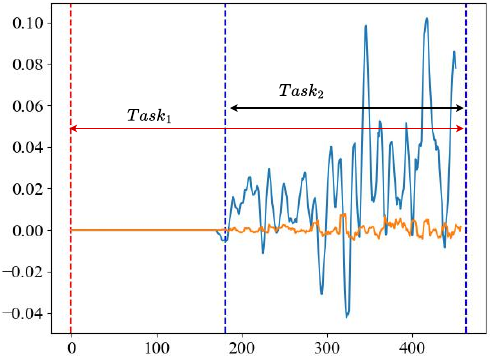}}
\subfigure{\centering
  \includegraphics[width=0.24\textwidth]{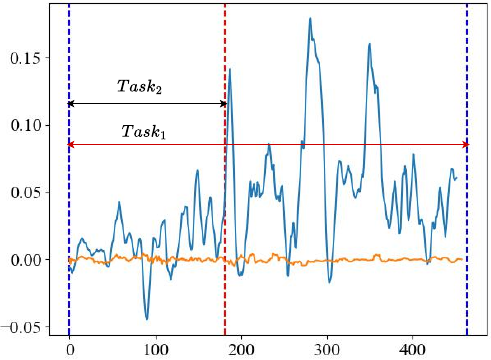}}
\subfigure{\centering
  \includegraphics[width=0.24\textwidth]{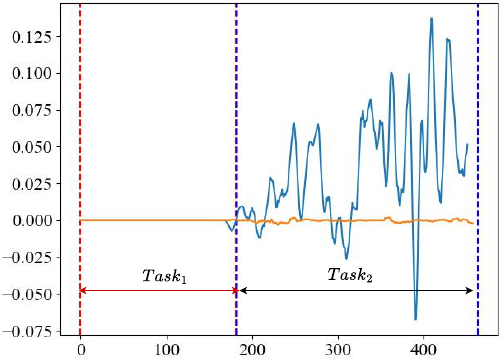}}
\caption{The variation in the gradient conflicts of the original gradient matrix $\mb{G}$ and adjusted gradient matrix $\hat{\mb{G}}$ under different timeline conditions.}
\label{fig:cos_sim}
\end{figure}

\begin{figure}[t]
\centering
\subfigure{\centering
  \includegraphics[width=0.257\textwidth]{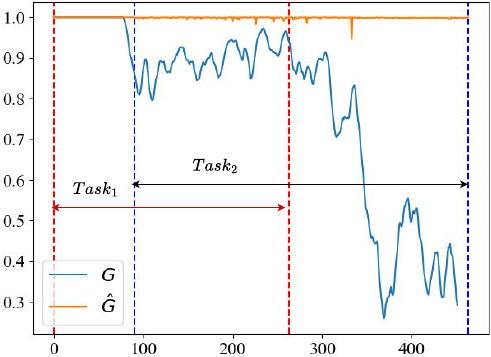}}
\subfigure{\centering
  \includegraphics[width=0.24\textwidth]{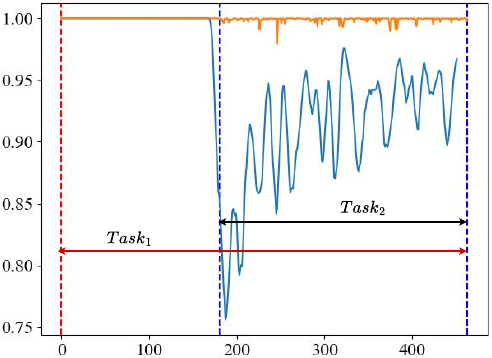}}
\subfigure{\centering
  \includegraphics[width=0.24\textwidth]{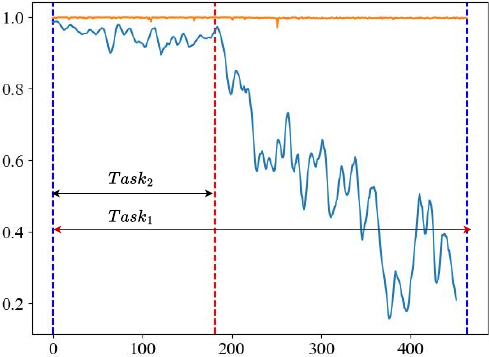}}
\subfigure{\centering
  \includegraphics[width=0.24\textwidth]{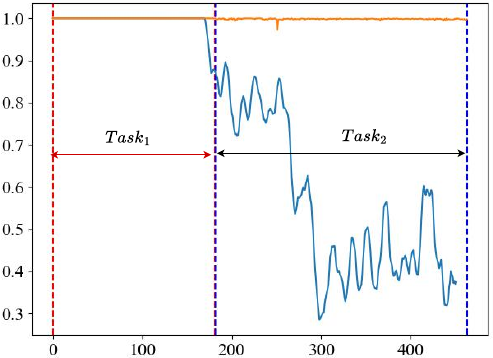}}
\caption{The variation in the gradient magnitude similarity of the original gradient matrix $\mb{G}$ and adjusted gradient matrix $\hat{\mb{G}}$ under different timeline conditions.}
\label{fig:mag_sim}
\end{figure}

\subsection{Hyperparameter selection}
\label{sec:hyper}

In Fig.~\ref{fig:hparms}, we show the major hyperparameter selection for regularization term $\alpha$ in Eq.~\eqref{eq:orth_loss} and regularization term $\sigma$, $\lambda$ in Eq.~\eqref{eq:block_descent}. The value of $\sigma$ effectively supports orthogonality enforcement and computational efficiency without leading to issues such as gradient instability or overly slow convergence. As $\sigma$ increases, its influence on the optimization objective gradually diminishes, eventually degrading to the standard orthogonalization condition $\hat{\mb{G}}^\top \hat{\mb{G}} = \mb{I}$, where the dashed line in Fig.~\ref{fig:hparms} represents the standard orthogonalization condition.
While increasing the regularization parameter $\sigma$ beyond a certain point does not significantly impact the overall effectiveness, we opt to set $\sigma = 100$ to maintain a balance between enforcing structural constraints and ensuring numerical stability and convergence speed.
The value of $\lambda$ corresponds to the constraint on the $\hat{\mb{G}}$ condition number. As $\lambda$ increases, the results exhibit an initial increase followed by a decrease. We select the optimal value at $\lambda = 100$.

\subsection{Reherasal analysis of SPCL}

We evaluate PCL with different sizes of rehearsal coreset on PS-CIFAR-100 datasets. The results are shown in Fig.~\ref{fig:mem}.
It shows that for both SORO and SORO+DBT$_{orth}$ methods, performance initially increases with the growth of the buffer size, then peaks, and starts to decline as more samples are stored. This trend indicates that while increasing the rehearsal buffer size initially helps in reducing catastrophic forgetting in PCL, it leads to diminishing returns beyond a certain point. Notably, SORO+DBT$_{orth}$ maintains a slight edge over SORO alone, suggesting that the DBT orthogonality constraints help stabilize learning and manage interference more effectively, even as buffer size increases.

\subsection{Toy experiments}

In this section, we assess the stability of training processes using the condition number, gradient conflicts, and gradient magnitude similarity as key stability metrics, as mentioned in Eq.~\eqref{eq:condition_number}.
Each experiment consists of a toy setup involving two tasks trained concurrently under PCL model conditions. The experimental setup varies in that the start and end times of the two tasks differ across trials. We compare the condition numbers of the original, unmodified gradient matrix $\mb{G}$ and the adjusted gradient matrix $\hat{\mb{G}}$, calculated using our proposed method, under various timeline conditions. The results, detailed in Fig.~\ref{fig:conds_2}, Fig.~\ref{fig:cos_sim} and Fig.~\ref{fig:mag_sim}, confirm that the gradient system becomes more stable with the application of our method, facilitating more effective and reliable training in dynamic continual learning environments. 

\newpage

\newpage

\end{document}